\pdfoutput=1
\documentclass{article}

\PassOptionsToPackage{numbers, sort, compress}{natbib}
\usepackage[preprint]{neurips_2025}
\makeatletter\renewcommand{\@notice}{}\makeatother
\usepackage[font=small]{caption}

\usepackage[utf8]{inputenc}
\usepackage[T1]{fontenc}
\usepackage{url}
\usepackage{booktabs}
\usepackage{amsfonts}
\usepackage{nicefrac}
\usepackage{microtype}
\usepackage{xcolor}
\usepackage{algorithm}
\usepackage{algpseudocode}
\renewcommand{\paragraph}[1]{\vspace{.1em}\noindent\textbf{#1}}

\usepackage{amsmath}
\usepackage{amssymb}

\definecolor{linkblue}{HTML}{6870ae}
\usepackage[pagebackref=true,colorlinks,citecolor=brown]{hyperref}
\hypersetup{
  urlcolor  = linkblue
}
\usepackage{multirow} 
\usepackage{makecell} 
\usepackage{wrapfig}
\usepackage{needspace}
\usepackage{svg}
\usepackage{cleveref}
\usepackage{caption}
\usepackage{fontawesome5}
\newcommand{\EBRlink}[3]{{\color{linkblue}#1\ \href{#2}{\textbf{#3}}}}
\usepackage[table]{xcolor}

\makeatletter
\newcommand{\wrapend}{\par
  \ifnum\c@WF@wrappedlines>\@ne
    \vspace*{\dimexpr\baselineskip*\numexpr\c@WF@wrappedlines-\@ne\relax\relax}%
  \fi
  \WFclear}
\makeatother

\usepackage{graphicx}
\usepackage{float}
\usepackage{placeins}
\floatstyle{plaintop}
\restylefloat{table}
\makeatletter
\g@addto@macro\fs@ruled{%
  \def\@fs@pre{\hrule height.8pt depth0pt \kern4pt}%
  \def\@fs@mid{\kern4pt\hrule\kern2pt}}
\makeatother
\usepackage[export]{adjustbox}

\usepackage{amsthm}
\usepackage{subcaption}
\usepackage{etoolbox}
\makeatletter
\patchcmd{\WF@floathand}{\ifvoid\z@\else\box\z@\fi}{\ifvoid\z@\hbox{}\else\box\z@\fi}{}
  {\PackageWarningNoLine{main}{Could not patch wrapfig; wrapped floats after \string\noindent\space may vanish}}
\makeatother
\usepackage{adjustbox}

\usepackage{pifont}

\usepackage{fontawesome5}

\algrenewcommand{\algorithmicrequire}{\textbf{Input:}}

\NewDocumentCommand{\heng}
{ mO{} }{\textcolor{red}{\textsuperscript{\textit{Heng}}\textsf{\textbf{\small[#1]}}}}

\NewDocumentCommand{\alexi}
{ mO{} }{\textcolor{red}{\textsuperscript{\textit{Alexi}}\textsf{\textbf{\small[#1]}}}}

\title{Energy-Based Reasoners: Your Recurrent Reasoner Should Be an Energy-Based Model}

\author{
  \textbf{William Xingxu Chen}$^{1}$\quad
  \textbf{Alexi Gladstone}$^{2}$\quad
  \textbf{Zeyu Michael Li}$^{1}$\quad
  \textbf{Heng Ji}$^{2}$\quad \textbf{Yilun Du}$^{3}$\\
  \normalfont $^{1}$Duke University\quad $^{2}$UIUC\quad $^{3}$Harvard University\\[4pt]
  \EBRlink{\faGlobe}{https://energy-based-reasoners.github.io}{energy-based-reasoners.github.io}\qquad
  \EBRlink{\faGithub}{https://github.com/wwillchen/ebr}{github.com/wwillchen/ebr}
}

\begin{document}

\maketitle

\begin{abstract}
\label{sec:abstract}
Looping is a powerful inductive bias for reasoning. By repeatedly iterating a recurrent block, looped models can scale their effective depth and enable multi-step reasoning in latent space. However, existing looped models are not inherently guided by an explicit measure of progress, making it difficult to determine when to stop looping and how to improve upon the current state. To enable this, we introduce Energy-Based Reasoners (EBRs), which learn an energy function over latent reasoning states and formulate looped updates as gradient descent on this energy. The energy scores latent states, while its gradient guides refinement and provides a signal for stopping. To train EBRs effectively, we further introduce exploration-based training to discover additional modes and to better shape the energy landscape over a broader set of states. Across four algorithmic reasoning benchmarks, EBRs match or outperform prior recurrent reasoning models, achieving 99.9\% accuracy on Sudoku-Extreme while using 345$\times$ less inference compute than the next-best model and maintaining comparable training compute. On all multi-solution benchmarks, EBRs achieve state-of-the-art coverage of distinct valid solutions while maintaining near-100\% single-sample accuracy. These results suggest that energy-based models are a promising approach for strong reasoning performance.
\end{abstract}
   
\section{Introduction}
\label{sec:intro}

\begin{figure}[t]
    \centering
    \includegraphics[width=0.9\textwidth]{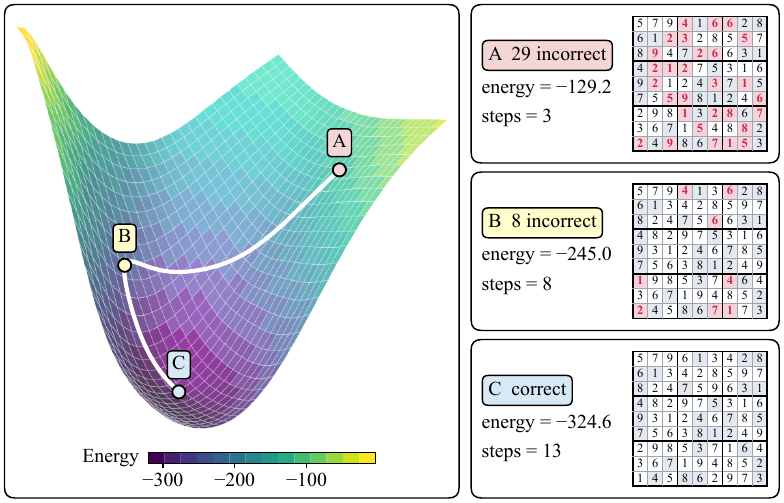}
    \caption{
    \textbf{Reasoning as energy minimization.}
    Energy-Based Reasoners (EBRs) solve a puzzle $x$ by gradient descent on a learned energy $E_\theta(z;x)$ over latent reasoning states $z$. The surface shows this energy for one Sudoku puzzle on the plane through three states of a descent trajectory.
    }
    \label{fig:energy_landscape}
\end{figure}

Generalization requires the ability to compute answers that cannot be retrieved. For a model, this means using what it has learned to construct solutions to unfamiliar problems~\citep{chollet2019measure}. For difficult reasoning tasks, this computation may involve a sequence of dependent steps, each building on the results of the last~\citep{saunshi2025reasoning}. Over the past few years, chain-of-thought has been widely used to support such multi-step reasoning in large language models. The model uses additional tokens to express intermediate reasoning steps before producing its final answer~\citep{wei2022chain}. In this setting, test-time scaling means generating more tokens to achieve better solutions, for example, generating longer reasoning traces, revising predictions, and searching over candidate solutions~\citep{snell2024scaling, wang2023selfconsistency, yao2023tree}.

A complementary approach carries out these reasoning steps through iterative updates to an internal latent state~\citep{hao2024training,geiping2025scaling}. Looped architectures organize these updates through recurrence~\citep{dehghani2018universal,giannou2023looped}. They repeatedly apply the same layers to a state that carries information from one reasoning step to the next. Reusing these layers allows the model to increase its computational depth without increasing its number of parameters~\citep{giannou2023looped,saunshi2025reasoning}. Recent works on looped reasoning models, including HRM and TRM, demonstrate the effectiveness of this approach on challenging algorithmic reasoning tasks, achieving strong performance with small models and sometimes outperforming much larger language models~\citep{wang2025hierarchical,jolicoeur2025less, gao2025universalreasoningmodel}.

However, moving reasoning into latent space raises the question of how we can actually guide updates to facilitate better search. Chain-of-thought solves this problem by exposing intermediate steps as token sequences that can be evaluated to decide which reasoning paths to pursue, how much computation to allocate to them, and which solution to select~\citep{cobbe2021training,lightman2023let,yao2023tree,snell2024scaling}. However, latent states do not directly expose a comparable explicit trace, and existing looped reasoners typically rely on separate mechanisms to evaluate predictions and determine stopping. For example, HRM and TRM use learned heads to discern between correct and incorrect states and decide when to halt~\citep{wang2025hierarchical,jolicoeur2025less,sghaier2026probabilistictinyrecursivemodel}, while FPRM uses convergence of the recurrent state as a stopping criterion~\citep{movahedi2026fixedpoint}. Other approaches introduce stochasticity to explore alternative latent trajectories and use additional criteria to select among the resulting predictions~\citep{huang2026equilibrium,sghaier2026probabilistictinyrecursivemodel,baek2026gram}. In each case, the evaluation is used only to stop or select, while the update itself is not guided by any explicit assessment of the current state. This motivates a broader question: \textit{can we learn a unifying objective so that evaluation guides the reasoning process, decides when to stop, and verifies which candidates are best?}

We argue that energy-based models offer a natural way to connect these capabilities. Prior work has demonstrated the potential of energy-based models for assessing uncertainty through learned energy scores~\citep{liu2020energy}, generating diverse samples via Langevin dynamics~\citep{du2019implicit}, and enabling iterative refinement through energy minimization~\citep{du2022learning,du2024learning,gladstone2025energy}. Building on these ideas, we introduce Energy-Based Reasoners (EBRs), which formulate reasoning as optimization over a learned energy landscape. Given a problem $x$, an EBR learns an energy function $E_\theta(z;x)$ over latent reasoning states $z$, assigning lower energy to states that produce better predictions. At inference time, reasoning proceeds by iteratively refining the latent state through gradient descent on this energy function. Importantly, the same learned energy also determines which predictions to select and when to stop refining them. Among multiple candidate solutions, an EBR selects the lowest-energy prediction without requiring a separate verifier. Meanwhile, the norm of the energy gradient provides a convergence signal, allowing each trajectory to stop once further refinement is no longer needed rather than relying on a fixed iteration budget. In this way, EBRs unify iterative refinement, candidate evaluation, and adaptive stopping within a single learned objective.

In order to generate diverse solutions, EBR must learn an energy landscape that captures multiple solution modes. Recent generative reasoning approaches, including Flow Reasoning Models and Looped Flows~\citep{helbling2026flowreasoningmodelsturning,suleymanzade2026thinkingloopedflows}, demonstrate the potential of generative modeling for recurrent reasoning. However, conventional generative frameworks, such as diffusion models, can suffer from exposure bias due to mismatches between training and inference. Explorative Modeling (XM)~\citep{gladstone2026explorativemodelingunlockingpretraining} offers an alternative by exploring model-generated candidates and matching them to suitable targets. This is particularly natural for energy-based reasoning, where supervising different trajectories provides learning signals across the latent states visited during descent, progressively shaping the energy landscape so that descent can better isolate modes. We therefore introduce state-based Explorative Modeling (SXM), extending XM to persistent energy-descent trajectories. By matching trajectories to solutions and supervising their intermediate states, SXM encourages the learned energy to guide refinement across a broader range of states and better realize the generative capacity of EBRs.

Across four challenging algorithmic reasoning benchmarks, Sudoku-Extreme, Maze-Hard, N-Queens, and Graph Coloring, EBRs match or outperform prior looped reasoning models. On Sudoku-Extreme, EBRs achieve 99.9\% accuracy while using 345$\times$ less inference compute than the next-best model, without requiring an external verifier. On multi-solution benchmarks, EBRs achieve high coverage of distinct valid solutions while maintaining near-perfect solution accuracy, demonstrating the ability of the learned energy landscape to support diverse reasoning outcomes. Together, these results establish energy minimization as an effective framework for accurate, efficient, and flexible iterative reasoning.

Our contributions are threefold,
\begin{itemize}
    \item We introduce Energy-Based Reasoners (EBRs), which unify iterative refinement, candidate evaluation, and adaptive stopping through a single learned energy function.
    \item We extend Explorative Modeling (XM) to latent reasoning trajectories through state-based Explorative Modeling (SXM), enabling the discovery of diverse solutions.
    \item We demonstrate improvements in reasoning accuracy, inference efficiency, and solution diversity across four algorithmic reasoning benchmarks.
\end{itemize}
\section{Related Work}
\label{sec:related_work}
\paragraph{Recurrent and recursive reasoners.}
Looped architectures perform step-by-step computation by repeatedly applying the same block of layers to a latent state that holds the model's current guess.
Each pass refines this guess, adding effective depth without additional parameters~\citep{dehghani2018universal,giannou2023looped,geiping2025scaling}. A recent line of work, HRM and TRM, has shown that small looped models can solve hard algorithmic reasoning problems much better than most large language models. They train the model to repeatedly refine a hierarchy of two latents.
Intuitively, the low-level latent acts as a scratchpad that is updated several times in quick succession, and the high-level latent, which holds the model's solution, is updated from the scratchpad~\citep{wang2025hierarchical,jolicoeur2025less}. However, both models are deterministic and can only generate one reasoning trajectory in inference, limiting their ability to explore alternative solutions.

\paragraph{Stochastic recursive reasoning.}
Several approaches introduce stochasticity into recursive reasoning
to explore different trajectories and candidate solutions. PTRM injects noise into a TRM-style recurrence and selects among the resulting candidates using the model's existing halting head~\citep{sghaier2026probabilistictinyrecursivemodel}.
GRAM instead approaches diversity from a generative modeling perspective, treating the reasoning trajectory as a latent variable trained with amortized variational inference so that independent samples recover different solutions~\citep{baek2026gram}.
Flow Reasoning Models and Looped Flows formulate recurrent refinement through learned denoising dynamics~\citep{helbling2026flowreasoningmodelsturning,suleymanzade2026thinkingloopedflows}. While these approaches demonstrate the potential of stochastic reasoning, stepwise generative training can introduce exposure bias. This motivates exploration-based supervision, where model-generated trajectories are matched to suitable targets. Our state-based Explorative Modeling (SXM) extends this idea to energy descent, providing supervision across different trajectories and intermediate states to better shape the learned energy landscape. 

\paragraph{Attractors and fixed-point reasoning.}
Recent studies have observed that recursive reasoning models can converge
to fixed points. Specifically, previous work has shown that HRM can
converge to both correct and incorrect fixed points, sometimes getting
stuck at incorrect solutions~\citep{ren2026reasoningmodelsreasoningguessing}.
EqR builds on this observation by learning task-conditioned attractors
and using damping and stochasticity to improve convergence toward
valid solutions~\citep{huang2026equilibrium}.
FPRM shows that fixed-point convergence can also be used to adapt
computation to problem difficulty. It introduces architectural
modifications and damping to stabilize long recurrences, allowing
a single latent state to be refined until convergence
~\citep{movahedi2026fixedpoint}.
Like FPRM, EBRs use a single latent state without hierarchy and
rely on convergence to determine when to stop. However, rather than learning an unconstrained recurrent update, EBR defines its dynamics through gradient descent on a learned energy function. This imposes structure on the update vector field and stationary points correspond to vanishing gradients. EBR further introduces convergence loss to encourage correct predictions to approach those stationary points. 

\paragraph{Energy-based reasoning.}
Energy-Based Models (EBMs) provide a natural formulation of fixed-point
reasoning that also supports candidate verification.
EBMs learn a scalar function that assigns an energy to a candidate
prediction given an input. The function is trained so that better
predictions receive lower energy~\citep{lecun2006tutorial}. Gradient descent on this energy gives an iterative update whose fixed points are exactly the stationary points of the energy, and the energy value itself acts as a learned score for comparing candidate solutions. Prior approaches differ in what the energy scores and how it is parameterized.
IREM and IRED apply this idea to algorithmic reasoning, with IREM
directly optimizing a learned energy over candidate solutions~\citep{du2022learning} and IRED incorporating diffusion-based
training and a sequence of annealed energy
landscapes~\citep{du2024learning}. Energy-Based Transformers similarly refine predictions by minimizing a learned energy~\citep{gladstone2025energy, davies2025ebt}, while Energy Transformers instead adopt an associative-memory formulation,
defining an energy over internal token representations and iteratively
refining these representations through energy
minimization~\citep{hoover2024energy}. Unlike IREM, IRED, and Energy-Based Transformers, EBRs use energy minimization to update a latent reasoning state rather than directly refining candidate predictions. And unlike Energy Transformers, EBRs do not use an associative-memory formulation; instead, they directly parameterize their scalar energy with a small transformer
trained end-to-end. Closer to recursive reasoning, ERM uses energies derived from known
task constraints to select among solutions generated by existing
reasoners~\citep{zhao2026energyguidedrecursivemodel}. Although ERM achieves strong results, its energy functions are designed around each task's rules rather than learned end-to-end, limiting the generality and scalability of the approach. In contrast, EBRs achieve stronger performance without relying on hard-coded task features. 

\begin{figure}[t]
\begin{minipage}[t]{0.485\textwidth}
\begin{algorithm}[H]
\caption{EBR training}
\label{alg:train}
\begin{algorithmic}[1]
\Statex \textbf{Input:} instance $x$; solution $y$
\Statex \textbf{Parameters:} segments $S$, noise scale $\sigma$,
        threshold $\delta$
\State $z\sim\mathcal{N}(0,I)$
\For{$s=1,\dots,S$}
    \State sample steps $n$ and step size $\alpha$
    \For{$t=1,\dots,n$}
        \State $\xi\sim\mathcal{N}(0,I)$
        \State $\tilde z\gets\operatorname{RMS}(z)+\sigma\xi$
        \State $z\gets\Pi_{\mathcal M}\bigl(\tilde z-\alpha\nabla_{\tilde z}E_\theta(\tilde z;x)\bigr)$
    \EndFor
    \State update $\theta$ with $\nabla_\theta\mathcal{L}$ \Comment{\cref{eq:loss}}
    \State \textbf{if} $r_n<\delta$ \textbf{then break} \Comment{\cref{eq:residual_halt}}
    \State $z\gets\operatorname{sg}(z)$
\EndFor
\end{algorithmic}
\end{algorithm}
\end{minipage}\hfill
\begin{minipage}[t]{0.485\textwidth}
\begin{algorithm}[H]
\caption{EBR inference}
\label{alg:infer}
\begin{algorithmic}[1]
\Statex \textbf{Input:} instance $x$
\Statex \textbf{Parameters:} segments $S$, steps $n$, step size $\alpha$,
        noise scale $\sigma$, threshold $\tau$
\State $z\sim\mathcal{N}(0,I)$
\setlength{\itemsep}{2.18pt}%
\For{$s=1,\dots,S$}
    \For{$t=1,\dots,n$}
        \State $\xi\sim\mathcal{N}(0,I)$
        \State $\tilde z\gets\operatorname{RMS}(z)+\sigma\xi$
        \State $z\gets\Pi_{\mathcal M}\bigl(\tilde z-\alpha\nabla_{\tilde z}E_\theta(\tilde z;x)\bigr)$
    \EndFor
    \State \textbf{if} $\lVert\nabla_{\mathcal{M}}E_\theta(z;x)\rVert<\tau$ \textbf{then break}
\EndFor
\State \Return $\widehat y_\theta(z)$
\end{algorithmic}
\end{algorithm}
\end{minipage}
\vspace{-8pt}
\end{figure}
\paragraph{Explorative modeling.}
Learning an energy landscape that supports multiple valid solutions
requires exploring and training on different reasoning trajectories.
Previous work has explored training with multiple predictions:
Multiple Choice Learning (MCL) and Best of Many (BoM) train using
the prediction or sample that best matches a
target~\citep{guzmanrivera2012multiple,bhattacharyya2018accurate},
while Implicit Maximum Likelihood Estimation (IMLE) matches each
training example to its closest generated
sample~\citep{li2018implicit,aghabozorgi2023adaptive,peng2022chimle}.
Explorative Modeling (XM) generalizes this matching principle to
both forward and reverse directions, applying it to end-to-end
generation as well as the training of existing generative
models~\citep{gladstone2026explorativemodelingunlockingpretraining}.
Forward XM selects the prediction that best matches a target,
while reverse XM selects the closest target to a generated prediction.
EBRs extend this bidirectional selection to recurrent reasoning
through state-based XM (SXM), selecting which latent trajectory
to supervise and which valid solution
best matches its prediction. 
By training on the states visited along different trajectories,
SXM expands coverage and shapes an energy landscape that better guides reasoning toward valid and diverse solutions. 
\section{Energy-Based Reasoning Model}
\label{sec:background}

\begin{figure}[t]
  \centering
  \includegraphics[width=0.9\textwidth]{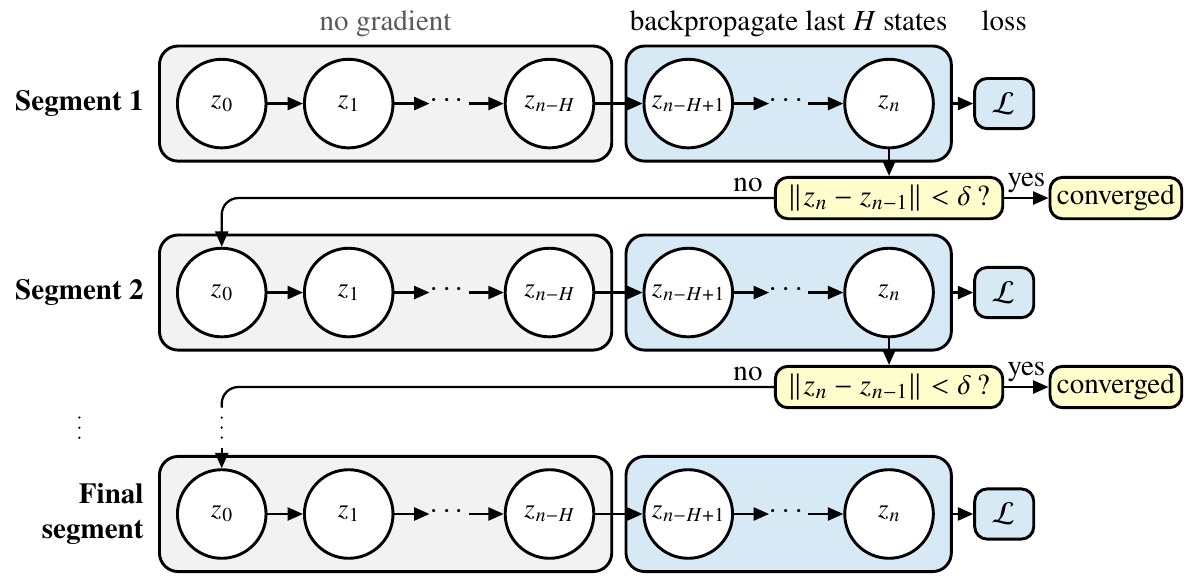}
  \caption{\textbf{Segment-wise training.}
  Each segment consists of $n$ descent steps.
  The first $n-H$ steps are computed without retaining a
  backpropagation graph (gray), while gradients propagate
  through the final $H$ steps (blue), whose states are
  supervised by $\mathcal L$ (\cref{eq:loss}). 
  After each segment, the loss is propagated and converged trajectories stop; otherwise, the detached endpoint initializes the next segment.}
  \label{fig:segment}
\end{figure}

\paragraph{Background: Looped reasoning models.}
We consider algorithmic reasoning problems $(x,y)$, where $x$ is a
problem instance and $y$ is a solution.
A recurrent reasoner predicts $y$ by iteratively refining a latent
state $z_t$ through a learned transition $f_\theta$,
\begin{equation}
    z_{t+1} = f_\theta(z_t,x),
    \qquad t=0,\ldots,T-1,
    \label{eq:transition}
\end{equation}
and decoding the final state $z_T$.
To train long recurrences, HRM~\cite{wang2025hierarchical}
and TRM~\cite{jolicoeur2025less} use \emph{deep supervision},
which divides the recurrence into segments, supervises their
predictions, and detaches the state between segments.
EBRs adopt this segment-wise training procedure, illustrated in
\cref{fig:segment}.

\paragraph{Energy-based reasoners.}
Rather than learning the transition $f_\theta$ directly, an EBR
learns a scalar energy $E_\theta(z;x)$ and updates its latent
state through gradient descent,
\begin{equation}
    z_{t+1} = z_t - \alpha \nabla_z E_\theta(z_t;x),
    \label{eq:ebr_update}
\end{equation}
where $\alpha$ is the step size.
The prediction $\widehat y_\theta(z_t)$ is decoded from the
current state $z_t$.
In practice, we use a variant of projected Langevin dynamics which injects
noise to encourage exploration and projects latent states onto
a sphere to stabilize descent and keep trajectories bounded.
See \cref{alg:infer} for the exact update. We denote this projection by $\Pi_{\mathcal{M}}$ and the
gradient tangent to the resulting manifold $\mathcal{M}$
by $\nabla_{\mathcal{M}}E_\theta$.

\subsection{Training Through Energy Descent}
\label{sec:training}

\paragraph{Segment-wise supervision.}
An EBR is trained on trajectories generated by its own energy
using deep supervision (\cref{fig:segment,alg:train}). Each segment $Z=(z_0,\ldots,z_n)$ contains $n$ descent steps. The initial state $z_0$ is sampled from noise or obtained by detaching the endpoint of the previous segment.
To bound memory usage, we use truncated backpropagation through time. This retains the gradients only through the final $H$ steps of each segment. The preceding $n-H$ steps are computed without gradient tracking. Since each update depends on the energy gradient, differentiating through the dynamics requires second-order derivatives of $E_\theta$. 

Let $\ell_\theta(z_t,y)$ denote the loss at state $z_t$, where $\ell_\theta(z_t,y)$ defines the stablemax cross-entropy loss following prior works~\cite{prieto2025grokkingedgenumericalstability, jolicoeur2025less, huang2026equilibrium}. The final $H$ states are supervised with,
\begin{equation}
    \mathcal{L}(Z,y)
    = \sum_{t=n-H+1}^{n} w_t\,\ell_\theta(z_t,y)
    + \lambda\,\mathcal{L}_{\mathrm{conv}}(Z,y),
    \label{eq:loss}
\end{equation}
where the weights $w_t$ increase toward the endpoint and
$\lambda$ controls the convergence loss.

\paragraph{Convergence supervision.}
Convergence alone does not imply correctness, as recurrent dynamics
can settle at fixed points that produce incorrect
predictions~\cite{huang2026equilibrium, ren2026reasoningmodelsreasoningguessing}.
We therefore introduce a convergence loss that encourages stationarity
to coincide with prediction correctness. We measure convergence using
the average relative change in latent states over the final $H$
steps of each segment,
\begin{equation}
    r_n = \frac{1}{H}\sum_{t=n-H+1}^{n}
    \frac{\|z_t-z_{t-1}\|}{\|z_{t-1}\|}.
    \label{eq:residual_halt}
\end{equation}
We then convert this residual into a convergence score
$p_n = \sigma((\delta-r_n)/\kappa)$, where $\delta$ is the
convergence threshold and $\kappa>0$ controls the sharpness
of the score. We supervise this score with the
correctness of the final prediction,
\begin{equation}
    \mathcal{L}_{\mathrm{conv}}(Z,y)
    = \mathrm{BCE}\bigl(
      p_n,
      \mathbf{1}[\widehat y_\theta(z_n)=y]
      \bigr).
    \label{eq:conv_loss}
\end{equation}
This encourages trajectories producing correct predictions to converge, while discouraging incorrect
predictions from becoming stationary. 
During training, trajectories halt when $r_n<\delta$
or when the maximum number of segments is reached.
Unlike a learned halt head, our convergence criterion is directly derived from the dynamics and requires no additional trainable parameters.

\paragraph{Randomized descent.}
To keep the learned energy from relying on a single descent schedule, each training segment samples its number of
descent steps $n$ and its step size $\alpha$, which stays fixed within the segment (\cref{alg:train}). Evaluation uses a
fixed number of steps and step size. The energy must therefore support refinement across depths and step lengths
rather than along one particular trajectory. We include the training details in \Cref{sec:energy_training}. 

\needspace{16\baselineskip}
\subsection{State-Based Explorative Modeling}
\label{sec:exploration}

\begin{wrapfigure}{r}{0.45\textwidth}
  \vspace{-\intextsep}
  \centering
  \includegraphics[width=\linewidth]{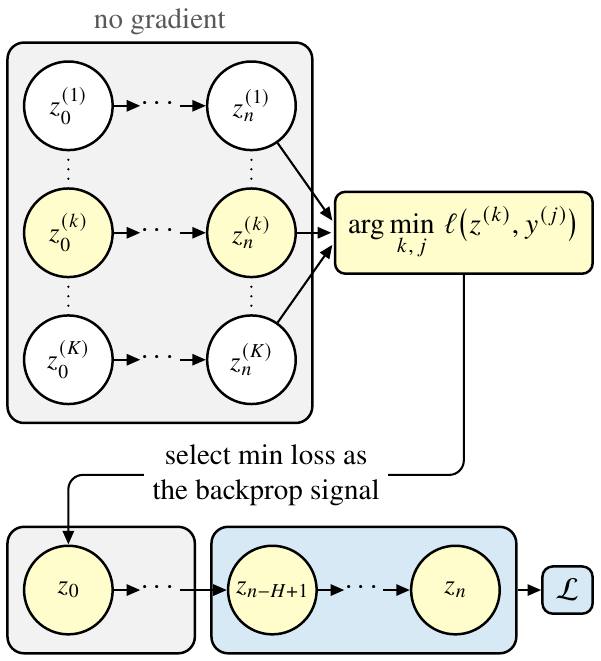}
  \caption{\textbf{State-based XM.}
  $K$ candidate trajectories are rolled out without
  gradients, and the trajectory--target pair with the
  lowest prediction loss is selected. The selected trajectory is replayed, with gradients
  propagated through its final $H$ steps
  (\cref{fig:segment}).}
  \label{fig:xm}
  \vspace{-\baselineskip}
\end{wrapfigure}

Training on a single trajectory only supervises the energy
along a limited region of the latent space, leaving other
regions of the energy landscape unexplored.
Moreover, when a problem admits multiple valid solutions,
supervising every trajectory against the same target can
discourage alternative correct predictions.
To address these limitations, we extend Explorative Modeling
(XM)~\cite{gladstone2026explorativemodelingunlockingpretraining}
from independent predictions to reasoning trajectories
based on energy minimization. Our state-based variant (SXM) maintains $K$
parallel descent trajectories per input across training segments.
At each segment, all $K$ trajectories are advanced, but only
the trajectory best matched to a valid solution is supervised.
Importantly, unselected trajectories continue evolving and remain
eligible for selection in later segments. Over $S$ segments,
SXM therefore explores up to $K\times S$ states while
supervising only one trajectory per segment. This allows
training to progressively shape the energy landscape along
different descent paths, while accommodating distinct modes that correspond to different solutions.\footnote{This progressive refinement resembles hierarchical exploration
methods such as CHIMLE~\cite{peng2022chimle} and Discrete Distribution Networks~\cite{yang2025ddn}. SXM differs from these previous works by exploring within the model's latent state rather than over independent candidate predictions.}

More concretely, let $z_n^{(k)}$ denote the endpoint of the
$k$-th trajectory at a given training segment. We consider
two directions for matching trajectories to targets.
\emph{Forward SXM (F-SXM)} selects the lowest-loss trajectory
for a given target $y$, whereas \emph{reverse SXM (R-SXM)}
selects the best-matching target from the available valid
solutions $\mathcal{Y}(x)$ for a given trajectory,
\begin{equation}
k_{\mathrm{F}}^\star(y)
    = \arg\min_{1\leq k\leq K}
    \ell_\theta(z_n^{(k)},y),
\qquad
y_{\mathrm{R}}^\star(k)
    = \arg\min_{y'\in\mathcal{Y}(x)}
    \ell_\theta(z_n^{(k)},y').
\label{eq:sxm_directions}
\end{equation}
When both the trajectory and target are free to vary,
we select the pair with the lowest prediction loss. In practice, we apply the two
directions sequentially, i.e., forward SXM selects the trajectory, and reverse SXM then selects its
target (\cref{sec:sxm_details}).
The set $\mathcal{Y}(x)$ also does not need to include all valid
solutions. We study the effect of training on incomplete solution sets
in \cref{sec:rsxm_set_size}. The matching procedure is illustrated
in \cref{fig:xm}.
\subsection{Inference and Prediction Selection}

At inference time, an EBR initializes its latent state from Gaussian
noise and refines it through energy descent. We introduce an early stopping criterion that terminates when the tangent gradient norm
$\lVert\nabla_{\mathcal M}E_\theta(z;x)\rVert$ falls below
a threshold $\tau$, or after $S$ segments of $n$ descent steps. In practice, both the gradient norm and the residual work as early stopping criteria, as shown in \cref{fig:descent}. The inference procedure is summarized in \cref{alg:infer}.

The energy function also provides a criterion for selecting
among candidates generated by independent trajectories. As shown in \cref{fig:descent},
the learned energy distinguishes solved from unsolved boards.
An EBR therefore selects the lowest-energy endpoint and returns
its decoded prediction.

\begin{table}[t]
    \centering
    \caption{
        \textbf{Accuracy on Sudoku-Extreme and Maze-Hard.}
        EBRs achieve strong single-sample accuracy
        on both benchmarks while using the fewest parameters.
        For multiple samples, the number of samples, N, is shown
        in parentheses. EBRs select the lowest-energy solution, while baseline results are taken from their papers and follow their respective evaluation protocols.
    }
    \label{tab:main_results}
    \small
    \begin{tabular}{lccc}
        \toprule
        Method & \# Params & Sudoku-Extreme (\textuparrow) & Maze-Hard (\textuparrow) \\
        \midrule
        \multicolumn{4}{c}{\textit{Single sample (\%)}} \\
        \midrule
        TRM~\citep{jolicoeur2025less}
            & 5--7M & 87.4 & 85.3 \\
        EqR~\citep{huang2026equilibrium}
            & 5--7M & 93.0 & -- \\
        FPRM~\citep{movahedi2026fixedpoint}
            & 7M & 94.2 & 87.0 \\
        GRAM~\citep{baek2026gram}
            & 10M & 86.6{\scriptsize$\pm$2.2} & -- \\
        Looped Flows~\citep{suleymanzade2026thinkingloopedflows}
            & 5--7M & 97.9{\scriptsize$\pm$0.4} & 86.7{\scriptsize$\pm$1.1} \\
        \rowcolor{blue!4}
        \textbf{EBR (Ours)}
            & \textbf{1.8M} & \textbf{99.4}{\scriptsize$\pm$0.1} & \textbf{87.4}{\scriptsize$\pm$1.1} \\
        \midrule
        \multicolumn{4}{c}{\textit{Multiple samples (\%)}} \\
        \midrule
        PTRM~\citep{sghaier2026probabilistictinyrecursivemodel}
            & 5--7M & 98.8 (100) & 86.7 (100) \\
        EqR~\citep{huang2026equilibrium}
            & 5--7M & 99.8 (128) & -- \\
        GRAM~\citep{baek2026gram}
            & 10M & 96.1{\scriptsize$\pm$0.9} (20) & -- \\
        Looped Flows~\citep{suleymanzade2026thinkingloopedflows}
            & 5--7M & 99.3{\scriptsize$\pm$0.2} (5) & 86.9{\scriptsize$\pm$1.5} (5) \\
        \rowcolor{blue!4}
        \textbf{EBR (Ours)}
            & \textbf{1.8M} & \textbf{99.9}{\scriptsize$\pm$0.1} (5) & \textbf{88.3}{\scriptsize$\pm$0.3} (5) \\
        \bottomrule
    \end{tabular}
\end{table}

\section{Experiments}
\label{sec:experimentation}

We evaluate EBRs on four algorithmic reasoning benchmarks to investigate three questions: (1) Can energy-based recurrence achieve strong reasoning performance and generalization from limited training data? (2) Can the learned energy landscape provide reliable signals for adaptive stopping and selecting among candidate solutions? (3) Can stochastic energy descent recover diverse valid solutions, and how does state-based explorative modeling (SXM) improve this capability?

To answer these questions, we benchmark Sudoku-Extreme and Maze-Hard~\citep{wang2025hierarchical,jolicoeur2025less}, evaluating reasoning accuracy and generalization capabilities of EBRs. We then investigate whether convergence of energy descent can guide adaptive stopping and whether learned energy values can reliably select among independently generated candidate solutions. Finally, we benchmark N-Queens and Graph Coloring~\citep{baek2026gram}, where multiple valid solutions exist, to examine the diversity of solutions recovered through stochastic descent and how SXM helps the learned energy landscape better capture these solutions. Our results are presented using mean $\pm$ standard deviation over three training seeds.

\subsection{Reasoning Performance and Test-Time Scaling}
\label{sec:reasoning_performance}

\begin{figure}[t]
  \centering
  \includegraphics[width=\textwidth]{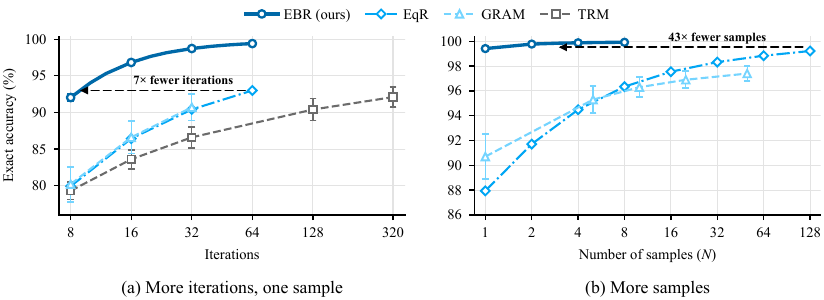}
  \caption{\textbf{Test-time scaling on Sudoku-Extreme.} Exact accuracy as we increase
  \textbf{(a)} the number of iterations with one sample and \textbf{(b)} the number of samples.
  EBRs select the lowest-energy sample. GRAM and TRM are from \citet{baek2026gram}.
  EqR is evaluated from its released checkpoint.}
  \label{fig:sudoku_depth}
\end{figure}

\begin{figure}[t]
    \centering
    \includegraphics[width=\linewidth]{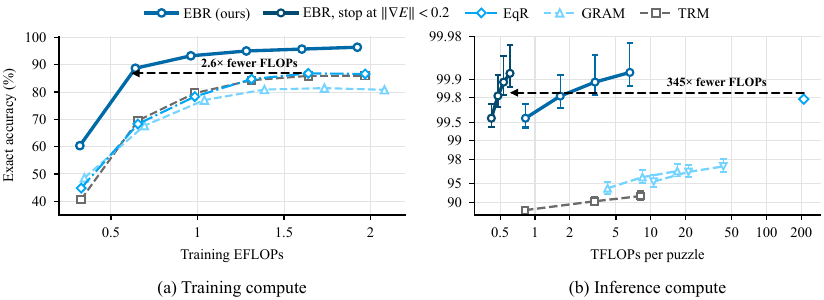}
    \caption{
        \textbf{Training and inference efficiency on Sudoku-Extreme.}
        \textbf{(a)} Single-sample exact accuracy at 16 iterations versus training compute.
        \textbf{(b)} Exact accuracy versus inference FLOPs per puzzle.
        We compare EBRs with and without convergence-based
        stopping against GRAM, EqR and TRM, using $N{=}8$ samples for EBRs, $N{=}20$ and $50$ for GRAM, $N{=}128$
        for EqR, and $N{=}1$ for TRM. GRAM and TRM are from \citet{baek2026gram}, and EqR is its reported
        accuracy~\citep{huang2026equilibrium}. Convergence-based stopping allows EBRs to maintain
        high accuracy with $10.8 \times$ less inference computation compared to EBRs with fixed iterations.
    }
    \label{fig:flops_frontier}
\end{figure}

\paragraph{Setup.}
We first evaluate whether energy minimization
supports accurate reasoning and generalization, and whether continued
iteration improves the model's predictions.
We consider Sudoku-Extreme and Maze-Hard~\citep{wang2025hierarchical}.
Sudoku-Extreme consists of $9\times9$ puzzles with unique solutions,
with 1,000 training puzzles augmented through validity-preserving
transformations and 422,786 test puzzles.
We report exact accuracy which requires all 81 cells to be correct.
Maze-Hard consists of $30\times30$ mazes, with 1,000 training and
1,000 test instances, where the task is to find a specific shortest path between a designated start and goal.

\paragraph{Benchmark performance.}
\Cref{tab:main_results} compares EBRs against existing recursive reasoning models on both benchmarks. All baseline results in \cref{tab:main_results} are taken from the original papers. With a single
sample, EBRs achieve 99.4\% on Sudoku-Extreme and 87.4\% on Maze-Hard, outperforming every compared method with
only 1.8M parameters. EBRs also show stronger multi-sample accuracy, achieving 99.9\% and 88.3\% with significantly fewer samples than previous state-of-the-art methods. \Cref{fig:sudoku_depth}b highlights this sample efficiency: EBRs approach peak accuracy with only a few samples, whereas competing methods improve more gradually. On Sudoku-Extreme, EBRs surpass EqR with just three samples, achieving 99.85\% accuracy compared with EqR's 99.8\% using 128 samples (\cref{fig:sudoku_depth}b). This sample efficiency translates into substantially lower inference FLOPs despite the additional cost of computing energy gradients at each recurrent step (\cref{fig:flops_frontier}b). EBRs also achieve higher accuracy at every evaluated training-compute budget (\cref{fig:flops_frontier}a).

\paragraph{Effect of continued refinement.}
EBRs also scale well with additional descent steps.  \Cref{fig:sudoku_depth}a shows that accuracy on Sudoku-Extreme steadily improves with increasing refinement depth, consistently outperforming the baselines at matched iteration budgets. Despite being trained with only four iterations, EBRs continue to improve their predictions through further descent on the same learned energy landscape, without requiring additional parameters, although the gains diminish with more iterations. 

\begin{figure}[t]
  \centering
  \includegraphics[width=\textwidth]{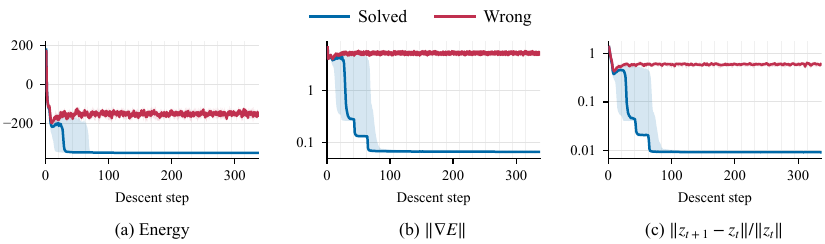}
  \caption{
    \textbf{Correct solutions converge to low-energy stationary states.}
    Energy (a), tangent gradient norm (b), and relative stepwise
    residual (c) throughout descent on Sudoku-Extreme, grouped
    by whether the final prediction is correct.
    Solved trajectories reach lower energies with small gradients
    and residuals, while incorrect trajectories continue to
    exhibit larger gradients and residuals.
  }
  \label{fig:descent}
\end{figure}

\needspace{6\baselineskip}
\subsection{Adaptive Computation and Solution Selection}
\label{sec:stop_select}

\setlength{\columnsep}{14pt}
\begin{wrapfigure}{r}{0.5\textwidth}
  \vspace{-\intextsep}
  \centering
  \includegraphics[width=\linewidth]{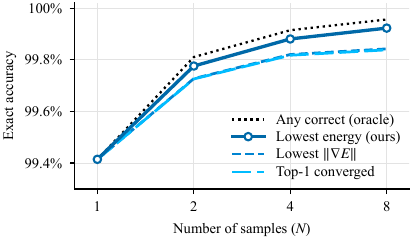}
\caption{
    \textbf{Energy-based selection closely approaches oracle accuracy.}
    Sudoku-Extreme exact accuracy when selecting one of $N$
    independently generated samples using the lowest energy,
    lowest gradient norm, or convergence criterion. Oracle uses test label.
}
  \label{fig:energy_selection}
\end{wrapfigure}
\paragraph{Adaptive stopping.}
We next investigate whether convergence provides a reliable
signal for stopping early. The descent curves in \cref{fig:descent} show that correct
predictions approach stationary points, while incorrect predictions
tend to maintain larger gradients (\cref{fig:descent}b) and continue changing (\cref{fig:descent}c). This suggests that convergence can provide a useful signal
for determining when further iterations are needed. Based on this observation, EBRs stop each trajectory when the projected
energy-gradient norm falls below a threshold $\tau$.
With $\tau=0.2$, EBRs maintain 99.9\% exact accuracy on Sudoku-Extreme while using $10.8\times$ fewer inference FLOPs than a
fixed-iteration budget (\cref{fig:flops_frontier}b and \cref{tab:tau}). At comparable accuracy, EBRs require up to $345\times$
fewer FLOPs than EqR.

\paragraph{Energy-based solution selection.}
While small gradient norms or residuals can indicate convergence, they do not guarantee correctness, as trajectories
may also settle at spurious minima. The learned energy provides an additional signal, since solved trajectories end at
lower energy than incorrect ones (\cref{fig:descent}a). Selecting the lowest-energy endpoint among independent
candidates therefore picks the correct answer more often than selecting the most converged one, and comes close to
oracle selection (\cref{fig:energy_selection}).
\wrapend

\subsection{Multi-Solution Reasoning}
\label{sec:multi_solution}

\begin{table}[t]
    \centering
    \caption{
        \textbf{Evaluation on N-Queens and Graph Coloring.}
        $N{=}1$ is a single sample. For $N{=}20$, EBRs keep the lowest-energy of 20 samples, and the baselines use majority vote.
        Cov@20 is the fraction of distinct valid solutions found with 20 samples per input.
        Conflict is the number of violated edges summed over the test graphs. Baseline results are from \citet{baek2026gram}.\protect\footnotemark
    }
    \label{tab:nqueens_graphcolor}

    \small
    \setlength{\tabcolsep}{2.5pt}
    \renewcommand{\arraystretch}{1.10}

    \begin{adjustbox}{max width=0.98\textwidth}
    \begin{tabular}{l c c cccc cccc}
        \toprule
        & & &
        \multicolumn{4}{c}{\textbf{N-Queens}}
        & \multicolumn{4}{c}{\textbf{Graph Coloring}} \\
        \cmidrule(lr){4-7}
        \cmidrule(lr){8-11}
        & & &
        \multicolumn{2}{c}{$8\times8$}
        & \multicolumn{2}{c}{$10\times10$}
        & \multicolumn{2}{c}{8-vertex}
        & \multicolumn{2}{c}{10-vertex} \\
        \cmidrule(lr){4-5}
        \cmidrule(lr){6-7}
        \cmidrule(lr){8-9}
        \cmidrule(lr){10-11}
        \textbf{Method}
        & \textbf{\# Params}
        & $\boldsymbol{N}$
        & \textbf{Acc.}$\uparrow$
        & \textbf{Cov@20}$\uparrow$
        & \textbf{Acc.}$\uparrow$
        & \textbf{Cov@20}$\uparrow$
        & \textbf{Conflict}$\downarrow$
        & \textbf{Cov@20}$\uparrow$
        & \textbf{Conflict}$\downarrow$
        & \textbf{Cov@20}$\uparrow$ \\
        \midrule

        Looped TF & 7M & --
        & 68.1{\scriptsize$\pm$7.4} & 23.6{\scriptsize$\pm$1.9}
        & 50.0{\scriptsize$\pm$7.6} & 6.2{\scriptsize$\pm$3.2}
        & 136.0{\scriptsize$\pm$16.1} & 20.5{\scriptsize$\pm$1.5}
        & 157.3{\scriptsize$\pm$9.0} & 7.2{\scriptsize$\pm$0.7} \\

        HRM & 27M & --
        & 78.7{\scriptsize$\pm$2.9} & 26.7{\scriptsize$\pm$1.3}
        & 37.8{\scriptsize$\pm$0.7} & 4.7{\scriptsize$\pm$0.1}
        & 109.7{\scriptsize$\pm$1.5} & 21.8{\scriptsize$\pm$0.3}
        & 164.3{\scriptsize$\pm$21.6} & 8.9{\scriptsize$\pm$1.7} \\

        TRM & 7M & --
        & 66.8{\scriptsize$\pm$5.6} & 22.6{\scriptsize$\pm$0.8}
        & 17.5{\scriptsize$\pm$11.2} & 2.0{\scriptsize$\pm$1.3}
        & 109.3{\scriptsize$\pm$3.1} & 22.3{\scriptsize$\pm$0.6}
        & 170.7{\scriptsize$\pm$17.9} & 6.8{\scriptsize$\pm$0.3} \\

        \midrule
        \multirow{2}{*}{GRAM} & \multirow{2}{*}{10M} & 1
        & 99.5{\scriptsize$\pm$0.3} & --
        & 89.7{\scriptsize$\pm$2.7} & --
        & 8.3{\scriptsize$\pm$2.5} & --
        & 31.3{\scriptsize$\pm$15.0} & -- \\
        & & 20
        & 100.0{\scriptsize$\pm$0.0} & 89.7{\scriptsize$\pm$1.7}
        & \textbf{100.0}{\scriptsize$\pm$0.0} & 57.5{\scriptsize$\pm$3.4}
        & 2.7{\scriptsize$\pm$2.1} & 85.8{\scriptsize$\pm$0.5}
        & 3.3{\scriptsize$\pm$1.5} & 51.3{\scriptsize$\pm$2.8} \\

        \midrule
        \rowcolor{blue!4}
        & & 1
        & \textbf{100}{\scriptsize$\pm$0} & --
        & \textbf{99.51}{\scriptsize$\pm$0.5} & --
        & \textbf{0}{\scriptsize$\pm$0} & --
        & \textbf{0}{\scriptsize$\pm$0} & -- \\
        \rowcolor{blue!4}
        \multirow{-2}{*}{\textbf{EBR (Ours)}} & \multirow{-2}{*}{\textbf{7M}} & 20
        & \textbf{100}{\scriptsize$\pm$0} & \textbf{98.1}{\scriptsize$\pm$0.2}
        & 99.99{\scriptsize$\pm$0.01} & \textbf{92.0}{\scriptsize$\pm$0.9}
        & \textbf{0}{\scriptsize$\pm$0} & \textbf{88.6}{\scriptsize$\pm$0.1}
        & \textbf{0}{\scriptsize$\pm$0} & \textbf{54.5}{\scriptsize$\pm$0.5} \\
        \bottomrule
    \end{tabular}
    \end{adjustbox}
\end{table}

\paragraph{Setup.}
We evaluate the ability of EBRs to generate multiple valid solutions
on the N-Queens and Graph Coloring benchmarks of
GRAM~\citep{baek2026gram}.
N-Queens provides partial boards with 5--7 queens removed in
the $8\times8$ setting and 7--9 removed in the $10\times10$ setting.
Graph Coloring uses three-colorable Erd\H{o}s--R\'enyi graphs
with eight or ten vertices.
For each input, we draw 20 independent samples and measure their validity and their coverage of distinct valid
solutions (\cref{tab:nqueens_graphcolor}).

\paragraph{Validity and diversity.}
Across all four benchmarks, EBRs achieve near-perfect validity while recovering more diverse solutions than prior methods.\footnotetext{GRAM's code caps coverage at 20 solutions per input (the most 20 samples can find) on N-Queens
but not on Graph Coloring. Applying the intended cap, our Graph Coloring coverage is 89.0\% and 85.3\%, which we
encourage future work to report (\cref{sec:evaluation_details}).} With a single sample, EBRs reach 100\% and 99.51\% accuracy on $8\times8$ and $10\times10$ N-Queens, respectively, and zero conflicts on both Graph Coloring tasks, outperforming GRAM even with 20 samples on the latter. With 20 samples, EBRs maintain near-perfect validity while achieving higher coverage of distinct valid solutions across all four benchmarks. The improvement is particularly significant on $10\times10$ N-Queens, where coverage increases from 57.5\% for GRAM to 92.0\% for EBRs.

\subsection{Ablations and Analysis}
\label{sec:ablations}
\setlength{\columnsep}{14pt}
\begin{wrapfigure}[24]{r}{0.5\textwidth}
    \vspace{\dimexpr-\intextsep-1pt\relax}
    \centering
    \captionsetup{type=table,position=top}
    \captionof{table}{\textbf{Sudoku-Extreme training ablations.} Each row changes one component of the EBR ($K{=}1$); the bottom rows add SXM with $K$ candidates. q-head selects smallest last step.}
    \label{tab:sudoku_ablations}
    \small
    \setlength{\tabcolsep}{4pt}
    \begin{tabular}{lcc}
        \toprule
        Variant & N=1$\uparrow$ & N=5 $\uparrow$ \\
        \midrule
        EBR ($K{=}1$) & 98.96 & 99.79 \\
        $-$ energy head & 60.05 & 76.59 \\
        $-$ randomized $\alpha$ & 97.59 & 99.39 \\
        $-$ convergence loss & 45.32 & 63.12 \\
        convergence loss $\to$ q-head & 92.42 & 95.56 \\
        \midrule 
        $+$ SXM ($K{=}2$) & \textbf{99.36} & \textbf{99.90} \\
        $+$ SXM ($K{=}3$) & 99.17 & 99.78 \\
        $+$ SXM ($K{=}4$) & 99.27 & 99.70 \\
        \bottomrule
    \end{tabular}

    \vspace{4pt}
    \captionsetup{type=table,position=top}
    \captionof{table}{\textbf{SXM ablations} on $8{\times}8$ N-Queens and 8-vertex Graph Coloring ($N{=}1$). Each row adds to the one above, and the last row is the model of \cref{tab:nqueens_graphcolor}.}
    \label{tab:xm_ablations}
    \small
    \setlength{\tabcolsep}{4pt}
    \begin{tabular}{lcccc}
        \toprule
        & \multicolumn{2}{c}{NQ $8{\times}8$} & \multicolumn{2}{c}{GC 8-vertex} \\
        \cmidrule(lr){2-3} \cmidrule(lr){4-5}
        Variant & Acc.$\uparrow$ & Cov.$\uparrow$ & Conflict$\downarrow$ & Cov.$\uparrow$ \\
        \midrule
        EBR (no XM) & 65.40 & 75.9 & 18 & 83.4 \\
        $+$ F-SXM ($K{=}2$) & 72.18 & 87.5 & 3 & 86.1 \\
        $+$ R-SXM & \textbf{100} & \textbf{97.9} & \textbf{0} & \textbf{88.5} \\
        \bottomrule
    \end{tabular}
\end{wrapfigure}
\paragraph{Energy-based recurrence.}
To isolate the role of energy descent, we replace EBR's gradient-based updates with a directly learned recurrent transition (\cref{tab:sudoku_ablations}). This reduces single-sample accuracy on Sudoku-Extreme from 98.96\% to 60.05\%, demonstrating the importance of energy-based recurrence. We note that removing the energy head does not reproduce TRM's \cite{sghaier2026probabilistictinyrecursivemodel, jolicoeur2025less} settings as it uses a smaller model (1.7M versus 5--7M parameters) and retains EBR's training setup, including a single latent state, self-attention, noise and normalization at each update, and convergence-based halting (\cref{sec:arch_details}). By changing only the recurrence mechanism, this comparison isolates the contribution of energy descent within EBR's framework. 

Randomizing the step size exposes the model to a broader range of descent trajectories and latent states, improving generalization. Fixing $\alpha$ reduces single-sample accuracy from 98.96\% to 97.59\%. Convergence supervision is quite important for preventing overfitting in the absence of another halting mechanism like a learned halting head (q-head). Removing it drops accuracy to 45.32\%.  Replacing the convergence loss with q-head, as in HRM and TRM~\citep{wang2025hierarchical,jolicoeur2025less}, partially recovers the accuracy drop.

\paragraph{State-based explorative modeling.}
We also include other training ablations and show that training with convergence loss performs better than training with a q-head loss. Training with SXM ($K=2$) improves single-sample accuracy, and larger $K$ brings no further gain. However, accuracy with five samples remains comparable, as energy-based selection already
recovers correct solutions for nearly all instances. In addition to Sudoku, we also examine the contributions of forward and reverse SXM on N-Queens and Graph Coloring (\cref{tab:xm_ablations}).
Each improves both validity and coverage, and we see that combining both improves accuracy and coverage on both benchmarks. 

\par\WFclear
% If Section 5's heading cannot fit, break with \vfil so \flushbottom does not stretch the previous page's gaps.
\needspace{5\baselineskip}
\section{Limitations and Future Work}
\label{sec:future_work}

While EBRs achieve strong performance, energy-based recurrence introduces additional computational overhead. Training through energy descent requires second-order derivatives, making each update more expensive than a directly learned recurrent transition. SXM further increases training cost by rolling out $K$ candidate trajectories per training segment. We measured this overhead by comparing training FLOPs to previous models in \cref{sec:training_flops}, and on Sudoku-Extreme EBRs are more accurate than the baselines at every training budget (\cref{fig:flops_frontier}a). To address these computational requirements, we optimized our implementation for computing second-order gradients, which required additional engineering such as custom GPU kernels (which are provided in our codebase). Nevertheless, we demonstrate that the accuracy gains justify the additional compute and obtain better results using less inference compute (\cref{fig:flops_frontier}b). Hence, developing more compute efficient energy-based models is an important direction for future work.

Training energy-based models also introduces additional hyperparameters, such as the step size, the noise scale and the
number of descent steps (\cref{sec:energy_training,tab:settings}). Unlike prior
energy-based models, which are often sensitive to these choices, we found them broadly transferable across benchmarks. Still, these additional hyperparameters add further complexity in configuring and training EBRs. 

We also note that SXM, and XM in general, tends to benefit from larger training sets. On Maze-Hard, SXM brought no gain, so we train it with $K=1$ (\cref{tab:settings}). Maze-Hard provides only 1,000 training mazes without augmentation, whereas Sudoku-Extreme augments its 1,000 training puzzles, resulting in roughly one million training boards. With little data, the additional trajectories explored by SXM have few distinct examples to cover, and we believe this limits both SXM and the overall gains on Maze-Hard (\cref{sec:maze_limitations}). 

Our experiments are also currently limited to structured algorithmic reasoning tasks, leaving open whether energy-based recurrence can scale to larger models and more general reasoning problems. With the recent renewed interest in looped architectures, an important direction is exploring whether learned energies can guide recurrent computation in tasks such as reasoning problems for language modeling or planning in world models.  

\section{Conclusion}
\label{sec:conclusion}

We introduced Energy-Based Reasoners (EBRs), reasoning
models that reason by gradient descent on a learned energy function.
By formulating reasoning as energy minimization, EBRs learn a single function to guide refinement, halting, and selection among candidate solutions. We further introduced state-based explorative modeling (SXM), which trains the energy on diverse reasoning trajectories
to support multiple valid solutions.
Across four algorithmic reasoning benchmarks, EBRs match or outperform prior looped models. With only 1.8M parameters, EBRs achieve 99.4\% single-sample accuracy on Sudoku-Extreme and 87.4\% on Maze-Hard. On Sudoku-Extreme, EBRs further achieve 99.9\% accuracy while using 345$\times$ less inference compute than the next-best model. On N-Queens and Graph Coloring, EBRs maintain near-perfect single-sample accuracy while recovering more distinct valid solutions than prior methods, raising coverage on $10\times10$ N-Queens from 57.5\% to 92.0\%. These results suggest that learning an energy landscape provides a promising alternative to directly learning recurrent state transitions for algorithmic reasoning.

\section{Acknowledgement}
Thanks to Purvish Jajal for ideas regarding XM and reasoning models. 
A.G. is supported by the National Science Foundation Graduate Research Fellowship Program under Grant No. DGE 21-46756. Any opinions, findings, and conclusions or recommendations expressed in this material are those of the author(s) and do not necessarily reflect the views of the National Science Foundation.  This research used the Delta and DeltaAI advanced computing and data resources, which are supported by the National Science Foundation (award OAC 2320345 and award OAC 2005572) and the State of Illinois. Delta and
DeltaAI are joint efforts of the University of Illinois Urbana-Champaign and its National Center for Supercomputing Applications. We also acknowledge the Duke Compute Cluster (DCC), maintained by Duke Research Computing, for providing computational resources used in this work.
{
    \small
    \bibliographystyle{unsrtnat}
    \bibliography{references}
}

\clearpage
\appendix

\renewcommand{\thesection}{\Alph{section}}
\renewcommand{\thefigure}{\Alph{section}\arabic{figure}}
\renewcommand{\thetable}{\Alph{section}\arabic{table}}

\setcounter{section}{0}
\counterwithin{figure}{section}
\counterwithin{table}{section}

\clearpage
\section{Model and Training Details}
\label{sec:experimental_details}

\subsection{Energy Network}
\label{sec:arch_details}

The energy $E_\theta$ reuses the recurrent Transformer block
of TRM and EqR~\citep{jolicoeur2025less, huang2026equilibrium} as its backbone and adds a
lightweight per-token head that reduces the block's output
to a scalar. \Cref{tab:trm_vs_ebr} summarizes the key
differences between the two models.

\begin{table}[ht]
    \centering
    \caption{
        \textbf{Key mechanism differences between TRM and EBRs.}
    }
    \label{tab:trm_vs_ebr}
    \small
    \begin{tabular}{@{}l>{\raggedright\arraybackslash}p{0.33\textwidth}>{\raggedright\arraybackslash}p{0.38\textwidth}@{}}
        \toprule
        & TRM & EBR \\
        \midrule
        Block output & the next latent state & a scalar energy (\cref{eq:energy_form}) \\
        State update & replaced by the block output & a step along $-\nabla_z E_\theta$ \\
        Latent state & hierarchical, $z_H$ and $z_L$ & a single $z$ \\
        Initial state & one fixed vector for every input & fresh draw from a normal \\
        Halting & learned Q head & residual of the descent (\cref{eq:residual_halt}) \\
        \bottomrule
    \end{tabular}
\end{table}

\paragraph{Backbone.}
The backbone consists of two Transformer blocks, each with
four-head self-attention and a SwiGLU MLP~\citep{shazeer2020glu},
followed by residual connections and RMSNorm~\citep{zhang2019root}.
We use self-attention across all tasks, whereas TRM replaces
attention with a token-axis MLP on Sudoku-Extreme.
The hidden dimension $d$ varies by task (\cref{tab:settings}).

As in TRM and HRM~\citep{wang2025hierarchical}, problem tokens
are embedded, scaled by $\sqrt{d}$, and added to the latent
state, so the backbone computes
$h=f_\theta(z+\operatorname{emb}(x))$.
We use two-dimensional rotary positional embeddings
(RoPE)~\citep{su2024roformer}, assigning half of each attention
head's channels to row positions and half to column positions.
Sudoku, Maze, and N-Queens use their board coordinates, while
Graph Coloring is treated as a one-dimensional sequence.

The key difference between EBRs and prior works lies in how the backbone is used.
TRM directly predicts the next latent state, whereas an EBR
uses the backbone to define a scalar energy and updates
the state along $-\nabla_z E_\theta$.
Each descent step therefore requires a forward and backward
pass through the backbone, and training through these steps
involves second-order derivatives.

Following TRM, linear layers are initialized from a truncated normal distribution and use no bias, except the hidden layer of the energy head (\cref{eq:energy_form}). We additionally learn
a separate attention temperature for each head, parameterized
in log space.

\paragraph{Energy head.}
\label{sec:energy_param}
The energy has the same form on every task,
\begin{equation}
    E_\theta(z;x)=\sum_{i=1}^{N}\phi(e_i),\qquad
    e_i=w^\top\operatorname{SiLU}\bigl(W h_i+b\bigr),\qquad
    h=f_\theta\bigl(z+\operatorname{emb}(x)\bigr),
    \label{eq:energy_form}
\end{equation}
where $z\in\mathbb{R}^{N\times d}$ holds one latent vector per input token, the head $(W,b,w)$ is shared by all tokens
and adds $d^2+2d$ parameters, and $\phi$ is the identity except on Maze-Hard, where it is a softplus. For example, the
head adds 66{,}048 parameters at $d=256$ (Sudoku-Extreme and Maze-Hard) and 263{,}168 at $d=512$ (N-Queens and Graph
Coloring), about 3.7\% of the model in both cases.
Although the energy is a sum of per-token terms, each
$h_i$ incorporates information from other tokens through
self-attention. Thus, interactions and constraints between
tokens are captured by the backbone, and the energy gradient
couples their updates during descent.

We find that the energy head can be adapted to different tasks through the choice of $\phi$. We use the identity function for all tasks except Maze-Hard, where we apply a softplus. With the identity, individual token contributions can be negative, and potentially favors solutions that contain more tokens of a particular type. This is less problematic for Sudoku, N-Queens, and Graph Coloring, where valid solutions have a fixed number of assigned positions. In Maze-Hard, however, valid paths can differ in length. We observe that the identity head learns negative contributions for path cells, allowing longer detours to receive lower energy than shorter paths. Using softplus constrains token energies to be nonnegative, and we observe that the model assigns higher contributions to path cells than to open cells, thereby discouraging unnecessary detours.

In addition to the energy network and token embeddings,
an EBR uses a linear readout to decode predictions from each
token's latent state. The full model has 1.8M parameters
at $d=256$ and 7.1M at $d=512$.

\subsection{Data and Settings}
\label{sec:training_details}

\paragraph{Data.}
We follow the original token layouts used by HRM for Sudoku-Extreme and Maze-Hard, and by GRAM for N-Queens and Graph Coloring. Sudoku-Extreme contains 1,000 training puzzles, each augmented with 1,000 validity-preserving transformations, while Maze-Hard contains 1,000 training mazes without augmentation.

For N-Queens and Graph Coloring, we use GRAM's datasets~\citep{baek2026gram}, preserving the original training and test splits. N-Queens instances are generated by enumerating all solutions for $N=8$ and $N=10$, then removing $k\in\{5,6,7\}$ and $k\in\{7,8,9\}$ queens, respectively. Distinct inputs are split 85:15 into training and test sets, yielding 4,376/772 examples for $N=8$ and 36,907/6,513 for $N=10$. Graph Coloring consists of three-colorable graphs with 8 or 10 vertices, with all valid colorings enumerated up to color permutation. The corresponding test sets contain 255 and 192 graphs, respectively. For reverse SXM, each training input is paired with its valid solutions.

\begin{table}[ht]
    \centering
    \caption{
        \textbf{Per-task settings.} Settings shared by all tasks are given in the text.
    }
    \label{tab:settings}
    \small
    \setlength{\tabcolsep}{5pt}
    \begin{tabular}{l cc cc cc}
        \toprule
        & & & \multicolumn{2}{c}{N-Queens} & \multicolumn{2}{c}{Graph Coloring} \\
        \cmidrule(lr){4-5} \cmidrule(lr){6-7}
        & Sudoku & Maze & $8{\times}8$ & $10{\times}10$ & 8-vertex & 10-vertex \\
        \midrule
        \multicolumn{7}{l}{\textit{Model}} \\
        Hidden size & 256 & 256 & 512 & 512 & 512 & 512 \\
        Parameters & 1.8M & 1.8M & 7.1M & 7.1M & 7.1M & 7.1M \\
        Per-cell energy & linear & softplus & linear & linear & linear & linear \\
        \midrule
        \multicolumn{7}{l}{\textit{Descent}} \\
        Noise scale $\sigma$ & 0.4$\to$0.05 & 0.4 & 0.3 & 0.3 & 0.3 & 0.3 \\
        Training segments $S$ & 4 & 6 & 3 & 3 & 4 & 4 \\
        Evaluation segments & 64 & 16 & 3 & 3 & 4 & 4 \\
        \midrule
        \multicolumn{7}{l}{\textit{Exploration}} \\
        Forward SXM candidates $K$ & 2 & 1 & 2 & 2 & 2 & 2 \\
        Reverse SXM & -- & -- & \checkmark & \checkmark & \checkmark & \checkmark \\
        Start from a noisy solution & -- & -- & \checkmark & \checkmark & \checkmark & \checkmark \\
        \midrule
        \multicolumn{7}{l}{\textit{Halting}} \\
        Loss weight $\lambda$ & 0.75 & 0.5 & 0.25 & 0.5 & 0.25 & 0.25 \\
        Threshold $\delta$ & 0.25 & 0.75 & 0.25 & 0.25 & 0.25 & 0.25 \\
        \midrule
        \multicolumn{7}{l}{\textit{Optimization}} \\
        Batch size & 768 & 64 & 768 & 768 & 768 & 768 \\
        Warmup steps & 100 & 500 & 1{,}000 & 1{,}000 & 1{,}000 & 1{,}000 \\
        Training steps & 60k & 50k & 30k & 45k & 70k & 70k \\
        \bottomrule
    \end{tabular}
\end{table}

\paragraph{Optimization.}
Hidden weight matrices are trained with Muon~\citep{jordan2024muon} (five Newton--Schulz iterations, momentum 0.95,
Nesterov). The learning rate of each matrix is scaled by 0.2 times the square root of its larger dimension, which
matches the update size of AdamW~\citep{liu2025muon}. Embeddings, the readout, the last layer of the energy head and
all vector parameters are trained with AdamW~\citep{loshchilov2017decoupled} ($\beta_1=0.9$, $\beta_2=0.95$). We use a base learning rate of $10^{-4}$ with linear warmup followed by a constant schedule, weight decay of 0.1. All evaluations use an exponential moving average of model weights with decay 0.999.

\paragraph{Compute.}
All experiments are trained on a single NVIDIA H200 GPU. \Cref{tab:compute} reports the training time for one seed, including intermediate evaluations.

\begin{table}[ht]
    \centering
    \caption{\textbf{Training time} of one seed on a single NVIDIA H200 GPU, including training evaluations.}
    \label{tab:compute}
    \small
    \setlength{\tabcolsep}{5pt}
    \begin{tabular}{l cc cc cc}
        \toprule
        & & & \multicolumn{2}{c}{N-Queens} & \multicolumn{2}{c}{Graph Coloring} \\
        \cmidrule(lr){4-5} \cmidrule(lr){6-7}
        & Sudoku & Maze & $8{\times}8$ & $10{\times}10$ & 8-vertex & 10-vertex \\
        \midrule
        Training steps & 60k & 50k & 30k & 45k & 70k & 70k \\
        Wall time (hours) & 3 & 3 & 4 & 8 & 5 & 8 \\
        \bottomrule
    \end{tabular}
\end{table}

\subsection{Training Through Energy Descent}
\label{sec:energy_training}

Training follows \cref{alg:train} (\cref{sec:training}). This section gives the settings of its randomized descent and
the remaining choices that stabilize energy descent and encourage the learned energy to support refinement beyond the
trajectories encountered during training.

\paragraph{Randomized number of descent steps.}
During training, we uniformly sample 15--21 descent steps per segment,
while evaluation uses a fixed 21 steps. For the first segment after a
trajectory reset, we instead sample 7--15 steps, exposing the model to
earlier stages of descent.

\paragraph{Randomized step size.}
We sample the step size independently for each trajectory and segment
as $\alpha\sim\mathcal{U}(200,250)$, keeping it fixed within the segment.
At evaluation, we use $\alpha=250$. Removing this randomization reduces
single-sample accuracy on Sudoku-Extreme (\cref{tab:sudoku_ablations}).

\paragraph{Noise and projection.}
We inject Gaussian noise before each descent step to encourage
exploration, with task-specific scales given in \cref{tab:settings}.
On Sudoku-Extreme, the noise scale follows a cosine schedule across
segments, decaying from 0.4 to 0.05; evaluation uses the same schedule.
Other tasks use a constant noise scale. To stabilize the dynamics,
we project latent states to unit RMS after each update using
$\Pi_{\mathcal M}$, corresponding to token-wise normalization onto
a sphere. Convergence residuals are computed on normalized states.

\paragraph{Truncated backpropagation and second-order gradients.}
To reduce training cost, we backpropagate only through the final
$H=7$ descent steps of each segment and compute earlier steps without
retaining a training computation graph. Since each descent update
depends on $\nabla_z E_\theta$, differentiating through these updates
requires computing a second-order derivative. The final state of
each segment is detached before initializing the next, preventing
gradients from propagating across segments.

\paragraph{Training objective.}
We supervise the final $H=7$ states of each segment with weights $w_t$ (\cref{eq:loss}) that increase linearly
toward later steps. The convergence
loss (\cref{eq:conv_loss}) computes the residual $r_n$ of \cref{eq:residual_halt} over the last 2 of the $H$
updates, with $\kappa=0.05$ and task-specific $\delta$
and $\lambda$ given in \cref{tab:settings}. Following HRM, with
probability 0.1 we delay halting until a randomly sampled number
of segments between 2 and $S$, encouraging training on longer
trajectories.

\paragraph{Initialization during training.}
While inference always starts from pure Gaussian noise, training on N-Queens and Graph Coloring also uses noisy solution embeddings, similar to the denoising supervision of IRED~\cite{du2024learning}. We randomly vary the amount of solution information in the initial state, ranging from pure noise to states that decode exactly to the target solution,
\begin{equation}
    z_0 = (I-P)\,\epsilon + \sqrt{1-a^2}\,P\epsilon + a\,s(y),
    \qquad \epsilon\sim\mathcal{N}(0,I),\quad a\sim\mathcal{U}(0,1),
    \label{eq:noisy_init}
\end{equation}
where $P$ projects each token's state onto the directions that the readout decodes, and $s(y)$ is a state that
decodes to the target solution $y$. Only the decodable part of the noise is mixed with the solution, so $a=0$
recovers the inference initialization and $a=1$ decodes exactly to $y$.
This encourages the learned energy to guide partially correct states toward valid solutions.

\paragraph{Robustness of energy training.}
Training through energy descent backpropagates through second-order derivatives of the energy, and this can cause gradient instability during training. We found gradient clipping to be helpful in addressing this. Not all benchmarks needed it, but we found gradient clipping particularly helpful on Maze-Hard.

\subsection{State-Based Explorative Modeling}
\label{sec:sxm_details}

Forward SXM rolls out $K$ candidates for each training row (\cref{tab:settings}). The candidates are ranked by
the cross-entropy of their endpoints against the stored solution of that row, the best one is replayed with gradients,
and all $K$ endpoints are carried into the next segment. On N-Queens and Graph Coloring, reverse SXM then regrades the
replayed trajectory. Each of its last three graded states, which include the endpoint, is compared with the valid
solutions of its input that appear in the same batch, and is graded against the one with the lowest cross-entropy.
In practice, more of the graded states, up to all $H$, can be regraded in the same way, but we only tried the last three.
The solution chosen for the endpoint is also the target of $\mathcal{L}_{\mathrm{conv}}$.

Because reverse SXM grades each trajectory against its closest valid solution, the trajectories of one input can all
be pulled toward the same solution, causing collapse. Following the reverse XM of
\citet{gladstone2026explorativemodelingunlockingpretraining}, we avoid this by not reusing solutions. Once a solution
has been assigned to one trajectory, the other trajectories in the training batch are graded against the
closest solution that has not been assigned yet. After 75\% of the input's solutions have been assigned, all of them
become available again.

\section{Training Compute Compared to GRAM}
\label{sec:training_flops}

\Cref{tab:training_flops} compares the training FLOPs and single-sample results of EBRs and GRAM~\citep{baek2026gram}. We measure the forward and backward computation of each training step, including second-order gradients through the descent steps of EBRs and the $K$ candidate rollouts used by SXM. With PyTorch's FLOP counter, we count matrix multiplications and attention products at two FLOPs per multiply-add, excluding elementwise operations, normalization, and optimizer updates. For EBRs, we use the configurations and training lengths in \cref{tab:settings}, averaging per-step costs over the randomized descent lengths. For GRAM, we use its training implementation and derive the total training steps from its dataset sizes, batch sizes, and number of epochs. Intermediate evaluations are excluded from both counts.

\begin{table}[ht]
    \centering
    \caption{
        \textbf{Training compute and single-sample results of EBRs and GRAM.} Total training FLOPs of the full run and
        the $N{=}1$ result, as in \cref{tab:main_results,tab:nqueens_graphcolor}. GRAM reports no Maze-Hard results.
    }
    \label{tab:training_flops}
    \small
    \setlength{\tabcolsep}{5pt}
    \begin{tabular}{l cc cc cc}
        \toprule
        & \multicolumn{2}{c}{Training steps} & \multicolumn{2}{c}{Total EFLOPs} & \multicolumn{2}{c}{$N{=}1$ result} \\
        \cmidrule(lr){2-3} \cmidrule(lr){4-5} \cmidrule(lr){6-7}
        Task & EBR & GRAM & EBR & GRAM & EBR & GRAM \\
        \midrule
        Sudoku-Extreme & 60k & 65.1k & 1.92 & 2.26 & 99.41{\scriptsize$\pm$0.12} & 86.6{\scriptsize$\pm$2.2} \\
        Maze-Hard & 50k & -- & 1.16 & -- & 87.37{\scriptsize$\pm$1.1} & -- \\
        N-Queens $8{\times}8$ & 30k & 47.0k & 2.45 & 1.08 & 100{\scriptsize$\pm$0} & 99.5{\scriptsize$\pm$0.3} \\
        N-Queens $10{\times}10$ & 45k & 140.8k & 5.82 & 4.74 & 99.51{\scriptsize$\pm$0.45} & 89.7{\scriptsize$\pm$2.7} \\
        Graph Coloring, 8-vertex & 70k & 45.5k & 2.46 & 0.67 & 0{\scriptsize$\pm$0} & 8.3{\scriptsize$\pm$2.5} \\
        Graph Coloring, 10-vertex & 70k & 87.6k & 3.98 & 1.77 & 0{\scriptsize$\pm$0} & 31.3{\scriptsize$\pm$15.0} \\
        \bottomrule
    \end{tabular}
\end{table}

On Sudoku-Extreme, EBRs use 15\% fewer training FLOPs than GRAM. On N-Queens and Graph Coloring, EBRs require 1.2--3.7$\times$ more total training FLOPs, primarily due to SXM candidate rollouts. In exchange, EBRs outperform GRAM on every shared task, with zero conflicts on both Graph Coloring tasks, and their coverage improvements are reported in \Cref{tab:nqueens_graphcolor}.

\paragraph{Evaluating EqR.}
Evaluating EqR at its full inference budget is expensive. Their inference configuration for their main results uses 128 samples and 64 iterations, so each Sudoku-Extreme puzzle costs about 210 TFLOPs, and the full test set of 422,786 puzzles would take 89 EFLOPs, or around 120 GPU-hours on an NVIDIA H200. We therefore compare against EqR's paper results in \cref{tab:main_results} and \cref{fig:flops_frontier}b: 93.0\% with one sample and 99.8\% with 128 samples~\citep{huang2026equilibrium}. For the sample-scaling curve in \cref{fig:sudoku_depth}b, we run their released checkpoint with EqR's own evaluation on Sudoku-Lite, the first 2,048 test puzzles. EqR's points in \cref{fig:sudoku_depth}b are therefore lower than its full-test results: 99.2\% with 128 samples and 87.9\% with one sample. However, the sample comparison in the same figure compares EBRs against EqR's reported accuracy of 99.8\%.
\FloatBarrier

\section{Evaluation Details}
\label{sec:evaluation_details}

\paragraph{Protocols.}
On Sudoku-Extreme, we evaluate all 422,786 test puzzles using eight independent samples, each run for up to 64 segments. On Maze-Hard, we evaluate 1,000 test mazes with five samples of 16 segments each. A maze is considered solved when its predicted path exactly matches the labeled solution; we additionally report any-shortest-path accuracy in \cref{sec:maze_limitations}. Pass@1 denotes the exact accuracy of the first sample, while Pass@$n$ selects the lowest-energy candidate among the first $n$ samples. Results are averaged over three training seeds. The ablations in \cref{tab:sudoku_ablations,tab:xm_ablations,tab:rsxm_set_size} use one seed.

\paragraph{N-Queens and Graph Coloring.}
We use GRAM's test splits~\citep{baek2026gram}. For each test input, we generate 20 independent samples, stopping each trajectory when its tangent gradient norm $\lVert\nabla_{\mathcal{M}}E_\theta\rVert$ falls below $\tau=0.2$, or when it reaches the maximum number of segments specified in \cref{tab:settings}. In \cref{tab:nqueens_graphcolor}, $N{=}1$ evaluates the first sample, while $N{=}20$ selects the EBR's lowest-energy sample. For N-Queens, a prediction is valid if it places $n$ non-attacking queens while preserving the given queens. For Graph Coloring, we report the total number of conflicting edges across test graphs. Ground-truth solutions are used only for evaluation and never for prediction selection.

\paragraph{Coverage.}
We generate 20 samples for each unique test input $x$. Let $\mathcal{S}(x)$ denote its solution set and
$\widehat y^{(1)},\ldots,\widehat y^{(20)}$ its decoded samples, and let
$\mathcal{F}(x)=\mathcal{S}(x)\cap\{\widehat y^{(1)},\ldots,\widehat y^{(20)}\}$ be the distinct solutions among
the samples. We define
\begin{equation}
    \mathrm{Cov@20}
    = \frac{\sum_{x}\lvert\mathcal{F}(x)\rvert}{\sum_{x}\min\bigl(\lvert\mathcal{S}(x)\rvert,20\bigr)},
    \label{eq:coverage}
\end{equation}
where the sums run over the unique test inputs. Since 20 samples can find at most 20 solutions, we cap the
denominator at 20 so that Cov@20 can reach 100\%. On N-Queens, $\mathcal{S}(x)$ contains the completions that keep
the given queens. On Graph Coloring, it contains the proper 3-colorings of $x$ up to color permutation. We use the
same definition on both tasks.

\paragraph{Differences from GRAM's code.}
GRAM's code~\citep{baek2026gram} computes coverage differently on the two tasks, and neither matches
\cref{eq:coverage} on inputs with more than 20 solutions. On Graph Coloring, it divides by
$\sum_x\lvert\mathcal{S}(x)\rvert$ without the cap. A graph with more than 20 colorings can then never be fully
covered, so no method can exceed 99.6\% coverage on 8-vertex graphs or 63.9\% on 10-vertex graphs. On N-Queens, it
caps the denominator at 20 but replaces $\mathcal{S}(x)$ by the first 20 solutions, so finding a solution outside
these 20 earns no credit. On N-Queens, the two definitions agree on inputs with at most 20 solutions, which covers
all $8\times8$ inputs and 6,494 of the 6,513 $10\times10$ inputs. On the remaining 19 inputs, GRAM's definition
lowers our $10\times10$ coverage from 92.0{\scriptsize$\pm$0.9}\% to 89.3{\scriptsize$\pm$1.0}\%. The GRAM results
in \cref{tab:nqueens_graphcolor} are taken from their paper and use their definitions. To match them, the table
reports our Graph Coloring coverage without the cap (88.6\% and 54.5\%). With the cap, it is
89.0{\scriptsize$\pm$0.1}\% and 85.3{\scriptsize$\pm$0.8}\%.

\paragraph{Inference compute.}
\Cref{tab:flops} reports the accuracy and per-iteration inference cost underlying \cref{fig:flops_frontier}b following the evaluation format of GRAM's Table~12~\citep{baek2026gram}.
\footnote{We thank the authors of GRAM for their extensive evaluation. We take the baseline and GRAM results directly from their Table~12~\citep{baek2026gram} and add our EBR evaluations to provide a consolidated reference for future work.}  FLOPs are counted per puzzle with PyTorch's FLOP counter. One iteration is one supervision step, and for an EBR it is one segment of 21 descent steps, each a forward and a
backward pass of the energy. With early stopping at $\lVert\nabla_{\mathcal{M}}E_\theta\rVert<0.2$, EBRs preserve the accuracy of the 16- and 64-iteration configurations while reducing their inference costs to 0.060{\scriptsize$\pm$0.002} and 0.076{\scriptsize$\pm$0.008} TFLOPs per puzzle, respectively. The five-sample cost is 0.380{\scriptsize$\pm$0.039} TFLOPs per puzzle. 

\begin{table}[ht]
    \centering
    \caption{
        \textbf{Accuracy (\%) on Sudoku-Extreme as a function of recursion iterations, with the per-iteration
        inference cost.} Baseline and GRAM rows are copied from
        GRAM's Table~12~\citep{baek2026gram}, where GRAM selects among its $N$ samples by majority vote. The EBR
        evaluates all 422{,}786 test puzzles, and selects the lowest-energy of its first $N$ samples. -- marks
        iteration counts that were not reported.
    }
    \label{tab:flops}
    \small
    \setlength{\tabcolsep}{4pt}
    \begin{adjustbox}{max width=\textwidth}
    \begin{tabular}{@{}lccccccc@{}}
        \toprule
        Model & TFLOPs / iter & 8 & 16 & 32 & 64 & 128 & 320 \\
        \midrule
        Looped TF & 0.0256 & 50.0{\scriptsize$\pm$2.5} & 61.7{\scriptsize$\pm$0.9} & 64.5{\scriptsize$\pm$0.9} & -- & 65.8{\scriptsize$\pm$1.0} & 66.1{\scriptsize$\pm$1.0} \\
        HRM & 0.0138 & 50.4{\scriptsize$\pm$1.1} & 56.8{\scriptsize$\pm$1.2} & 60.2{\scriptsize$\pm$1.3} & -- & 63.8{\scriptsize$\pm$1.5} & 65.4{\scriptsize$\pm$1.6} \\
        TRM & 0.0256 & 79.3{\scriptsize$\pm$1.2} & 83.6{\scriptsize$\pm$1.3} & 86.6{\scriptsize$\pm$1.4} & -- & 90.4{\scriptsize$\pm$1.5} & 92.1{\scriptsize$\pm$1.4} \\
        \midrule
        GRAM, $N{=}1$ & 0.0265 & 80.2{\scriptsize$\pm$2.4} & 86.6{\scriptsize$\pm$2.2} & 90.7{\scriptsize$\pm$1.8} & -- & -- & -- \\
        GRAM, $N{=}5$ & 0.1325 & 89.8{\scriptsize$\pm$2.0} & 93.6{\scriptsize$\pm$1.4} & 95.3{\scriptsize$\pm$1.1} & -- & -- & -- \\
        GRAM, $N{=}10$ & 0.2650 & 92.4{\scriptsize$\pm$1.6} & 95.1{\scriptsize$\pm$1.1} & 96.3{\scriptsize$\pm$0.8} & -- & -- & -- \\
        GRAM, $N{=}20$ & 0.5300 & 94.1{\scriptsize$\pm$1.3} & 96.1{\scriptsize$\pm$0.9} & 96.9{\scriptsize$\pm$0.7} & -- & -- & -- \\
        GRAM, $N{=}50$ & 1.3250 & 95.4{\scriptsize$\pm$1.0} & 96.8{\scriptsize$\pm$0.7} & 97.4{\scriptsize$\pm$0.6} & -- & -- & -- \\
        \midrule
        \textbf{EBR (Ours)}, $N{=}1$ & 0.0129 & 92.04{\scriptsize$\pm$0.43} & 96.83{\scriptsize$\pm$0.35} & 98.74{\scriptsize$\pm$0.21} & 99.41{\scriptsize$\pm$0.12} & -- & -- \\
        \textbf{EBR (Ours)}, $N{=}3$ & 0.0387 & 98.21{\scriptsize$\pm$0.35} & 99.37{\scriptsize$\pm$0.20} & 99.73{\scriptsize$\pm$0.12} & 99.85{\scriptsize$\pm$0.07} & -- & -- \\
        \textbf{EBR (Ours)}, $N{=}5$ & 0.0644 & 99.15{\scriptsize$\pm$0.26} & 99.69{\scriptsize$\pm$0.14} & 99.84{\scriptsize$\pm$0.09} & 99.90{\scriptsize$\pm$0.06} & -- & -- \\
        \textbf{EBR (Ours)}, $N{=}8$ & 0.1031 & 99.57{\scriptsize$\pm$0.18} & 99.81{\scriptsize$\pm$0.10} & 99.89{\scriptsize$\pm$0.07} & 99.92{\scriptsize$\pm$0.05} & -- & -- \\
        \bottomrule
    \end{tabular}
    \end{adjustbox}
\end{table}

\paragraph{Stopping threshold.}
\Cref{tab:tau} examines the trade-off between accuracy and inference computation as the stopping threshold $\tau$ varies on Sudoku-Extreme. Increasing $\tau$ from 0.1 to 1 reduces the average number of iterations from 7.6 to 3.5 while maintaining 99.4\% single-sample accuracy. At $\tau=0.2$, EBRs preserve both Pass@1 and Pass@8 accuracy while using $10.8\times$ fewer inference FLOPs than the fixed 64-iteration budget. More aggressive stopping reduces accuracy. At $\tau=2$, Pass@1 drops to 98.8\% while Pass@8 remains near 99.9\%, whereas $\tau=5$ causes a substantial degradation in both.

\begin{table}[ht]
    \centering
    \caption{
        \textbf{Effect of the stopping threshold $\tau$ on Sudoku-Extreme.} Iterations are averaged over samples, with a cap of 64. Pass@8 keeps the
        lowest-energy of eight samples. TFLOPs are per puzzle for eight samples and include the cost of the gradient
        check, and the saving is relative to running all 64 iterations.
    }
    \label{tab:tau}
    \small
    \begin{tabular}{@{}lccccc@{}}
        \toprule
        $\tau$ & Avg.\ iterations & Pass@1 & Pass@8 & TFLOPs / puzzle & Saving \\
        \midrule
        No stopping & 64 & 99.41{\scriptsize$\pm$0.12} & 99.92{\scriptsize$\pm$0.05} & 6.60 & 1.0$\times$ \\
        \midrule
        0.1 & 7.6{\scriptsize$\pm$1.4} & 99.41{\scriptsize$\pm$0.12} & 99.92{\scriptsize$\pm$0.05} & 0.82 & 8.0$\times$ \\
        0.2 & 5.6{\scriptsize$\pm$0.6} & 99.41{\scriptsize$\pm$0.12} & 99.92{\scriptsize$\pm$0.05} & 0.61 & 10.8$\times$ \\
        0.5 & 3.6{\scriptsize$\pm$0.1} & 99.42{\scriptsize$\pm$0.12} & 99.86{\scriptsize$\pm$0.07} & 0.39 & 17.1$\times$ \\
        1 & 3.5{\scriptsize$\pm$0.1} & 99.41{\scriptsize$\pm$0.12} & 99.88{\scriptsize$\pm$0.07} & 0.38 & 17.4$\times$ \\
        2 & 3.5{\scriptsize$\pm$0.1} & 98.81{\scriptsize$\pm$0.17} & 99.89{\scriptsize$\pm$0.06} & 0.38 & 17.6$\times$ \\
        5 & 2.1{\scriptsize$\pm$0.7} & 74.22{\scriptsize$\pm$14.49} & 95.66{\scriptsize$\pm$5.93} & 0.22 & 29.7$\times$ \\
        \bottomrule
    \end{tabular}
\end{table}

\section{Maze-Hard: Limitations and Shortest-Path Accuracy}
\label{sec:maze_limitations}

\paragraph{Limited gains.}
The improvements of EBRs on Maze-Hard are modest compared with Sudoku-Extreme. One possible explanation is the limited training data since the benchmark provides only 1,000 training mazes, which may be insufficient to learn the search behavior required by this task. However, we do not vary the training-set size, so whether additional data would improve performance remains an open question.

\begin{figure}[ht]
    \centering
    \includegraphics[width=\textwidth]{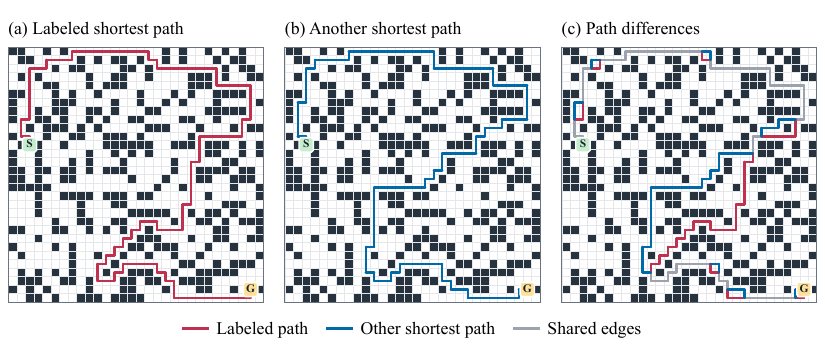}
    \caption{
        \textbf{Different shortest paths for the same Maze-Hard instance.}
        \textbf{(a)} The benchmark's labeled path and \textbf{(b)} an alternative both reach the goal (G) from the
        start (S) in 111 steps. \textbf{(c)} Path-specific edges are
        colored, and shared edges are gray. Exact match accepts only
        \textbf{(a)}, whereas shortest-path accuracy accepts both.
    }
    \label{fig:maze_shortest_paths}
\end{figure}

\paragraph{Exact match versus shortest paths.}
The original Maze-Hard evaluation requires predictions to match a single labeled shortest path, even when multiple shortest paths exist. The dataset labels can be reproduced exactly using breadth-first search with the direction priority $D<L<R<U$ for both training and test annotations. Consequently, an equally short alternative path is counted as incorrect (\cref{fig:maze_shortest_paths}). To distinguish failures in path finding from differences in path selection, we additionally report \emph{any-shortest-path accuracy}, which accepts any valid start-to-goal path whose length equals the shortest-path distance. 

Using the same checkpoints and trajectories as \cref{tab:main_results},
EBRs achieve 94.07\% single-sample shortest-path accuracy and
94.77\% with lowest-energy selection among five samples
(\cref{tab:maze_shortest_accuracy}). 

\begin{table}[ht]
    \centering
    \caption{
        \textbf{Exact-match and shortest-path accuracy (\%) on Maze-Hard.}
    }
    \label{tab:maze_shortest_accuracy}
    \small
    \begin{tabular}{@{}lcc@{}}
        \toprule
        Inference & Labeled-path exact match & Any shortest path \\
        \midrule
        EBR, Pass@1 & 87.37{\scriptsize$\pm$1.10} & 94.07{\scriptsize$\pm$0.91} \\
        EBR, Pass@5 & 88.27{\scriptsize$\pm$0.32} & 94.77{\scriptsize$\pm$0.59} \\
        \bottomrule
    \end{tabular}
\end{table}

\paragraph{Relation to prior work.}
EqR~\citep{huang2026equilibrium} introduces the revised Maze-Unique benchmark, which is also used by FRM~\citep{helbling2026flowreasoningmodelsturning}, making their reported maze results not directly comparable to ours. We retain the original benchmark, while reporting any-shortest-path accuracy to better reflect whether the predicted routes solve the underlying maze.

\section{Reverse SXM with Partial Solution Sets}
\label{sec:rsxm_set_size}

Reverse SXM matches each trajectory to the closest valid solution
in $\mathcal{Y}(x)$ (\cref{eq:sxm_directions}). Each training batch contains every
solution of its inputs, so $\mathcal{Y}(x)$ holds all valid solutions of $x$ in the training set
(\cref{sec:sxm_details}). To investigate whether
reverse SXM requires access to all valid solutions, we restrict
each training example to at most $k$ targets, retaining its
original solution and randomly selecting up to $k-1$ others. Evaluations are unaffected. 

\begin{figure}[ht]
    \centering
    \includegraphics[width=\textwidth]{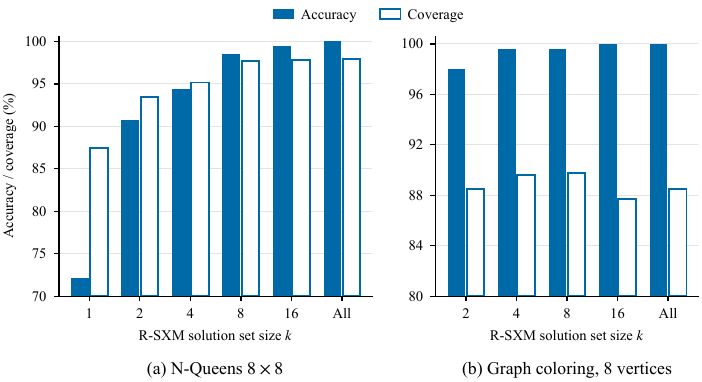}
    \caption{
        \textbf{Reverse SXM with partial solution sets.}
        We restrict reverse SXM to at most $k$ valid targets
        per training example and measure their accuracy and coverage. 
    }
    \label{fig:rsxm_set_size}

    \vspace{10pt}
    \captionsetup{type=table}
    \captionof{table}{
        \textbf{Values of \cref{fig:rsxm_set_size}} ($N{=}1$, in \%).}
    \label{tab:rsxm_set_size}
    \small
    \setlength{\tabcolsep}{8pt}
    \begin{tabular}{lcccc}
        \toprule
        & \multicolumn{2}{c}{N-Queens $8{\times}8$} & \multicolumn{2}{c}{Graph Coloring 8-vertex} \\
        \cmidrule(lr){2-3} \cmidrule(lr){4-5}
        $k$ & Acc.$\uparrow$ & Cov@20$\uparrow$ & Acc.$\uparrow$ & Cov@20$\uparrow$ \\
        \midrule
        1 & 72.18 & 87.5 & -- & -- \\
        2 & 90.73 & 93.5 & 98.04 & 88.5 \\
        4 & 94.35 & 95.2 & 99.61 & 89.6 \\
        8 & 98.55 & 97.7 & 99.61 & \textbf{89.8} \\
        16 & 99.52 & 97.8 & \textbf{100} & 87.7 \\
        All & \textbf{100} & \textbf{97.9} & \textbf{100} & 88.5 \\
        \bottomrule
    \end{tabular}
\end{figure}

\Cref{fig:rsxm_set_size} and \cref{tab:rsxm_set_size} show that a partial solution set already helps achieve strong accuracy and coverage. On $8\times8$ N-Queens, accuracy and coverage both increase with $k$:
from 72.18\% accuracy and 87.5\% coverage without reverse SXM ($k=1$) to
90.73\% and 93.5\% at $k=2$, 98.55\% and 97.7\% at $k=8$, and 100\% and
97.9\% with all solutions. On 8-vertex Graph Coloring, coverage stays
within about two points across $k$ (87.7--89.8\%), with comparable accuracy (98.0--100\%). Reverse SXM therefore does not require access to every valid
solution and performs well with partial sets. We did not tune hyperparameters separately for each $k$, so it remains unclear whether other values of $k$ could outperform the full solution set.

\end{document}